\documentclass[5pt]{elsarticle}

\usepackage{epsfig}
\usepackage{graphicx}
\usepackage{amsmath,amssymb} 
\usepackage{algorithm}
\usepackage{algorithmicx}
\usepackage{algpseudocode}
\usepackage{graphics}
\usepackage{threeparttable}
\usepackage{color}
\usepackage[normalem]{ulem}
\usepackage{multirow}
\usepackage{float}
\usepackage{amsfonts}
\usepackage{bm}
\usepackage{array}
\usepackage{pifont}
\usepackage{diagbox}
\usepackage{rotating}
\usepackage{booktabs}
\usepackage{overpic}
\usepackage{textcomp}
\usepackage{contour}
\usepackage{makecell}

\usepackage{enumitem}
\usepackage{colortbl}
\usepackage[american]{babel}
\usepackage{microtype}
\usepackage{bbding}

\usepackage{wrapfig} 
\usepackage{flushend} 

\graphicspath{{./Imgs/}{./Imgs/authors/}}
\DeclareGraphicsExtensions{.pdf,.jpg,.png}

\ifdefined \GramaCheck
\newcommand{\CheckRmv}[1]{}
\newcommand{\figref}[1]{Figure 1}%
\newcommand{\tabref}[1]{Table 1}%
\newcommand{\secref}[1]{Section 1}
\newcommand{\equref}[1]{Equation 1}
\newcommand{\algref}[1]{Algorithm 1}
\else
\newcommand{\CheckRmv}[1]{#1}
\newcommand{\figref}[1]{Fig.~\ref{#1}}
\newcommand{\tabref}[1]{Table~\ref{#1}}
\newcommand{\secref}[1]{Section~\ref{#1}}
\newcommand{\equref}[1]{Equ. (\ref{#1})}
\fi

\usepackage{colortbl}
\usepackage{cleveref}
\crefformat{section}{\S#2#1#3}
\crefformat{subsection}{\S#2#1#3}
\crefformat{subsubsection}{\S#2#1#3}

\usepackage{silence}
\def\sArt{{state-of-the-art~}}

\newcommand{\ourM}{{EMBANet}}

\usepackage{amssymb}

\date{}

\begin{document}

\begin{frontmatter}



\title{\ourM: A Flexible Efficient Multi-branch Attention Network}


\author[1,3,4]{Keke Zu\corref{equal1}}
\ead{zukeke@gmail.com}

\author[1,3]{Hu Zhang\corref{equal1}}
\ead{huzhang198@gmail.com}

\author[1,3]{Jian Lu\corref{cor1}}
\ead{jianlu@szu.edu.cn}

\author[2]{Lei Zhang \corref{cor1}}
\ead{cslzhang@comp.polyu.edu.uk}

\author[1,3]{Chen Xu}
\ead{chenxuszu@sina.com}

\address[1]{Shenzhen Key Laboratory of Advanced Machine Learning and Applications,  Shenzhen University, Shenzhen, China}
\address[2]{Department of Computing, The Hong Kong Polytechnic University, Hong Kong, China}
\address[3]{The National Center for Applied Mathematics Shenzhen (NCAMS), Shenzhen,  China}
\address[4]{Yangtze Delta Region Institute (Quzhou), University of Electronic Science and Technology of China, Quzhou, Zhejiang, China}
\cortext[equal1]{Keke Zu and Hu Zhang contributed equally.}
\cortext[cor1]{Jian Lu and Lei Zhang are corresponding authors.}


\begin{abstract}
This work presents a novel module, namely multi-branch concat (MBC), to process the input tensor and obtain the multi-scale feature map. The proposed MBC module brings new degrees of freedom (DoF) for the design of attention networks by allowing the type of transformation operators and the number of branches to be flexibly adjusted. Two important transformation operators, multiplex and split, are considered in this work, both of which can represent multi-scale features at a more granular level and increase the range of receptive fields. By integrating the MBC and attention module, a multi-branch attention (MBA) module is consequently developed to capture the channel-wise interaction of feature maps for establishing the long-range channel dependency. By substituting the 3x3 convolutions in the bottleneck blocks of the ResNet with the proposed MBA, a novel block namely efficient multi-branch attention (EMBA) is obtained, which can be easily plugged into the state-of-the-art backbone CNN models. Furthermore, a new backbone network called EMBANet is established by stacking the EMBA blocks. The proposed EMBANet is extensively evaluated on representative computer vision tasks including: classification, detection, and segmentation. And it demonstrates consistently superior performance over the popular backbones.

\end{abstract}




\end{frontmatter}



\section{Introduction}\label{sec:introduction}

The ability of the multi-scale feature representation is essential for various vision tasks,  such as image classification \cite{2019Res2Net, pyconv}, object detection \cite{DetectoRS, FPN, Salient_pyramid, 2018Pyramid}, instance segmentation \cite{deeplabv3,he2017mask}, semantic segmentation \cite{PAN, SANet_seg}, and scene parsing \cite{DAN_dualatt, pspnet}.
Multi-scale features are widely used in conventional network designs, as illustrated by
\cite{krizhevsky2012imagenet,simonyan2014very,szegedy2017inception,szegedy2015going,he2016deep,
huang2017densely,Chollet_2017_CVPR,xie2017aggregated,yu2018deep}. 
The efficiency of the multi-scale ability of convolutional neural networks (CNNs) can be subsequently improved in various ways, such as by utilizing features with different resolutions (e.g., \cite{chen2019drop,chen2018biglittle,cheng2019high,2020DyNet}), by using shortcut dense connections (e.g., DenseNet~\cite{huang2017densely}), or by using convolutional layers with different kernel sizes (e.g., InceptionNets~\cite{szegedy2017inception, szegedy2015going, szegedy2016rethinking}). In particular, the popular backbone Res2Net~\cite{2019Res2Net} uses the 3x3 convolution to connect channels in a cascading way, and generate a feature map that is rich in multi-scale spatial information. 

\begin{figure}[h]
\centering
\includegraphics[width=.8\linewidth]{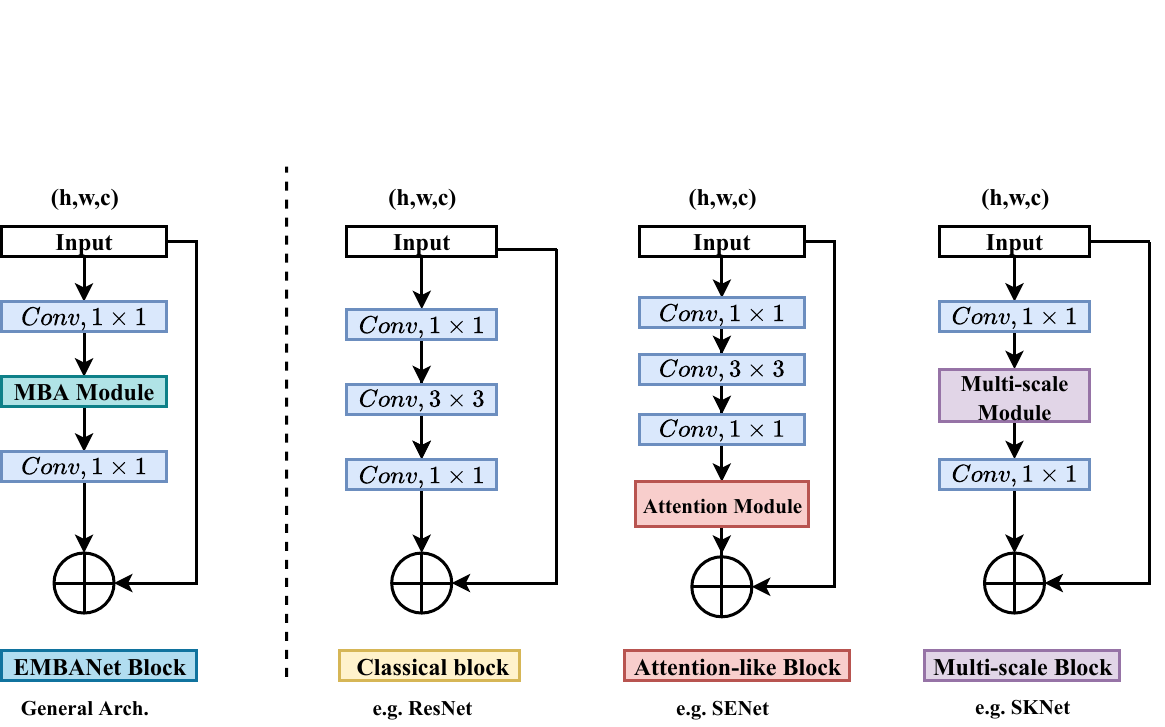}
\caption{The EMBANet block is presented as a unified framework for attention networks. By allowing the types of feature extraction operators and attention modules to be freely adjusted, the proposed EMBANet supports flexible network architectures. And thus, extra DoF is achieved by the proposed EMBANet.}

\label{fig:embanet-block}
\end{figure}

Differentiated from prior art, a novel module, namely Multi-branch and Concat (MBC), is proposed to process the input tensor. The proposed MBC-based architecture allows the type of transformation operators and the number of branches to be adaptively adjusted according to practical applications. There are various implementation choices on different types of transformation operators for MBC. In this work, we mainly focus on two important types namely Multiplex and Split, and thus two variations for implementing the MBC are developed: Multiplex and Concat (MUC) and Split and Concat (SPC). As illustrated by the two variations, the proposed MBC brings new \textbf{degrees of freedom} (DoF) to the design of attention networks by scaling or compressing the channel dimension of an input tensor and controlling the number of branches. Consequently, an effective module namely Multi-branch Attention (MBA) is developed by integrating the proposed MBC and classical attention module. As shown in Fig. \ref{fig:embanet-block}, a novel block namely Efficient Multi-branch Attention (EMBA) is obtained by substituting the 3x3 convolutions in the bottleneck blocks of the ResNet with the developed MBA module. Finally, a network called EMBANet is established by stacking these EMBA blocks, as illustrated by Fig. \ref{fig:embanet_arch}. The main contributions of this work are listed below:

\begin{itemize}
\item A novel module named Multi-branch and Concat (MBC) is proposed by allowing the type of transformation operators and the number of branches to be flexibly adjusted, which brings new \textbf{degrees of freedom} (DoF) to the design of attention networks.

\item Two important transformation operators, multiplex and split, are developed to implement the MBC.

\item A novel Multi-branch Attention (MBA) module, which can effectively extract multiscale spatial information at a more granular level and establish a long-range channel dependency, is developed based on the MBC module. 

\item Seven variations are demonstrated based on the proposed architecture of EMBANet and a large DoF is achieved.

\item Extensive experiments demonstrated that the proposed \ourM~achieves better performance on various computer vision tasks. 

\end{itemize}

\CheckRmv{
\begin{figure}[h]
\centering
\includegraphics[width=\linewidth]{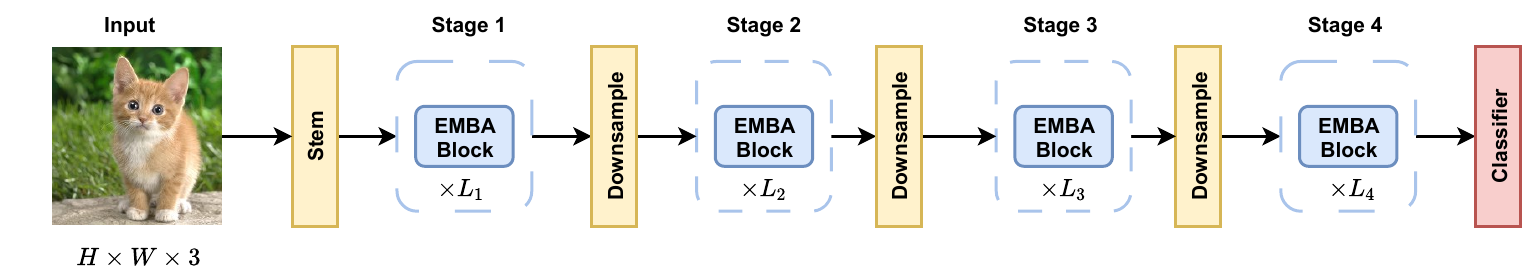}
\caption{The architecture of the proposed EMBANet, which has four hierarchical stages, each with a stack of EMBANet blocks and preceded by a downsampling operator. In particular, the stem unit consists of a 7$\times$7 convolution with stride 2 and a maxpooling operator. The quantity $L_{i}$ represents the number of EMBANet blocks at stage $i$, with the default set as (3, 4, 6, 3), and the classifier consists of an average pooling operator followed by a fully connected layer.}

\label{fig:embanet_arch}
\end{figure}
}

\section{Related Work}
\subsection{Attention Mechanism}
The attention mechanism is used to strengthen the allocation of the most informative feature expressions while suppressing the less useful ones, and thus makes the model attend to important regions within a context in an adaptive way. The Squeeze-and-Excitation (SE) attention in \cite{hu2018senet} can capture channel correlations by selectively modulating the scale of the channel. The CBAM in \cite{cbam} can enrich the attention map by adding max-pooled features for the channel attention with large-size kernels. Motivated by the CBAM, the GSoP in \cite{GSoP} proposes a second-order pooling method to extract richer feature aggregation. More recently, the Non-Local block \cite{Non-local_long} is proposed to build a dense spatial feature map and capture the long-range dependency via non-local operations. Based on the Non-Local block, the Double Attention Network($A^2$Net) \cite{doubleAAnet} introduces a novel relation function to embed attention with spatial information into the feature map. Sequentially, the SKNet in \cite{sknet} employs a dynamic selection attention mechanism that allows each neuron to adaptively adjust its receptive field size based on multiple scales of the input feature map. The ResNeSt \cite{zhang2020resnest} proposes a similar Split-Attention block that enables attention across groups of the feature map. The FcaNet \cite{2020FcaNet} proposes a novel multi-spectral channel attention mechanism that realizes the preprocessing in the frequency domain. The GCNet \cite{2019GCNet} introduces a simple spatial attention module, and thus a long-range channel dependency is established. The ECANet \cite{20ecanet} develops a one-dimensional convolution layer to reduce the redundancy of fully connected layers. The DANet \cite{DAN_dualatt} adaptively integrates local features with their global dependencies by summing these two attention modules from different branches. The methods discussed above either focus on the design of more sophisticated attention modules, which inevitably incur a higher computational cost, or they are incapable of establishing a long-range channel dependency. In this work, a novel module namely multi-branch attention (MBA) is proposed to further integrate the local and global attention information. And a more flexible architecture to design and implement attention networks is offered.

\subsection{Multi-scale Feature Representations}
The ability to represent multi-scale features is critical for various vision tasks and recent advances in CNN design have demonstrated that performance can be improved by a stronger multi-scale feature representation ability \cite{krizhevsky2012imagenet,simonyan2014very,szegedy2015going,he2016deep,huang2017densely,Chollet_2017_CVPR,xie2017aggregated,yu2018deep}. 

\subsubsection{Image Classification}
By introducing a 1$\times$1 convolution, the Network in Network (NIN)~\cite{lin2013network} inserts multi-layer perceptrons as micro-networks into the large network and the model discriminability of local patches is enhanced within the receptive field. By utilizing parallel filters with different kernel sizes, the GoogLeNet~\cite{szegedy2015going} can further enhance the multi-scale representation capability. By stacking more filters in each parallel path in the GoogLeNet, Inception networks~\cite{szegedy2016rethinking,szegedy2017inception} can further expand the receptive field. By employing the shortcut connection, the ResNet~\cite{he2016deep} and DenseNet~\cite{huang2017densely} can make objects processed on a very wide range of scales and their networks can really go deeper. The DPN~\cite{chen2017dual} inherits the advantages of ResNet and DenseNet, its hierarchical tree structure enables the network to obtain even stronger multi-scale representation abilities at individual layers. The DyNet\cite{2020DyNet} employs the dynamic convolution method to adaptively generate convolution kernels based on image content. The PyConv~\cite{pyconv} enlarges the receptive field by varying the size and depth of kernels at different levels. Besides, the PyConv processes the input tensor by using an increased kernel size in parallel to capture details of different levels. By constructing hierarchical residual-like connections within one single residual block, the Res2Net~\cite{2019Res2Net} represents multi-scale features at a more granular level and increases the range of receptive fields for each layer. The HRNet~\cite{SunXLW19,SunZJCXLMWLW19} introduces high-resolution representations and keeps the interaction of feature maps with different resolutions in a parallel way. 
Our preliminary work was implemented as the EPSANet~\cite{EPSANet} by using the Efficient Pyramid Squeeze Attention (EPSA) block. In this work, we are no longer limited to improving the specific attention model, but creatively propose a multi-branch architecture from the new perspective of DoF.  It is worth noting that the EPSANet can be considered as one of embodiments of the proposed EMBANet. 

\subsubsection{Image Segmentation}
The ESPNet~\cite{ESPNet} uses the spatial pyramid structure of dilated convolutions to re-sample the feature maps for learning representations from a large receptive field. The DeepLabv3~\cite{deeplabv3} applies several parallel atrous convolutions with different rates (called Atrous Spatial Pyramid Pooling, or ASPP), to capture the contextual information at multiple scales. The PSPNet~\cite{pspnet} exploits the capability of global context information by performing pyramid pooling operations at different grid scales, to incorporate suitable global features. 

\subsubsection{Object Detection}
The Feature Pyramid Network (FPN)~\cite{FPN} develops a top-down architecture with lateral connections to build high-level semantic feature maps at all scales. While, the DetectoRS~\cite{DetectoRS} uses a recursive feature pyramid module to incorporate extra feedback connections from FPNs into the bottom-up backbone layers. And the switchable atrous convolution is used to convolve the features with different atrous rates and gather results with switch functions.

Our proposed EMBA block enables the variation of receptive fields in channel dimensions to capture detailed information and global features. Thus, multi-scale features at a more granular level are extracted. 
The proposed EMBA block can be used as a plug-and-play block to be directly  integrated with CNNs and boost their performance.

\section{Method}
\subsection{Revisting Channel Attention}
The channel attention mechanism allows the network to selectively weight the importance of each channel and thus generate a more informative output. An SE block \cite{hu2018senet} consists of two parts: squeeze and excitation, which are respectively designed for encoding global information and recalibrating the channel-wise relationship adaptively. Generally, channel-wise statistics can be generated by using global average pooling, which is used to embed global spatial information into a channel descriptor. The input feature map is denoted by the quantity $X\in \mathbb{R}^{C\times H\times W}$, where $H$, $W$, $C$ represent the height, width, and the number of input channels respectively. The global average pooling operator can be calculated as
\begin{equation}
g_{c} = \frac{1}{H \times W} \sum\limits_{i=1}^H \sum\limits_{j=1}^W x_{c}(i,j)
\end{equation}
The attention weight of the $c$-th channel in the SE block can be written as
\begin{equation}
w_{c} = \sigma(W_{1} \delta (W_{0}(g_{c})))
\end{equation}
where the symbol $\delta $ represents the Rectified Linear Unit (ReLU) operation as in \cite{relu}, $W_{0}\in \mathbb{R}^{C\times\frac{C}{r}}$ and $W_{1}\in \mathbb{R}^{\frac{C}{r} \times C}$ are the fully-connected (FC) layers. With two FC layers are employed, the linear information among channels can be combined more efficiently. And it is helpful for the interaction of information from high and low channel dimensions. The symbol $\sigma$ represents the excitation function, and the Sigmoid function is commonly used in attention networks. With an excitation function, weights are assigned to channels after the channel interaction and thus the information can be extracted more efficiently. The above introduced process of generating channel attention weights is called the SE module in \cite{hu2018senet}, which is illustrated by Fig. \ref{fig:se_w_module}. 
\CheckRmv{
\begin{figure}[h]
\centering
\begin{overpic}[width=0.8\linewidth]{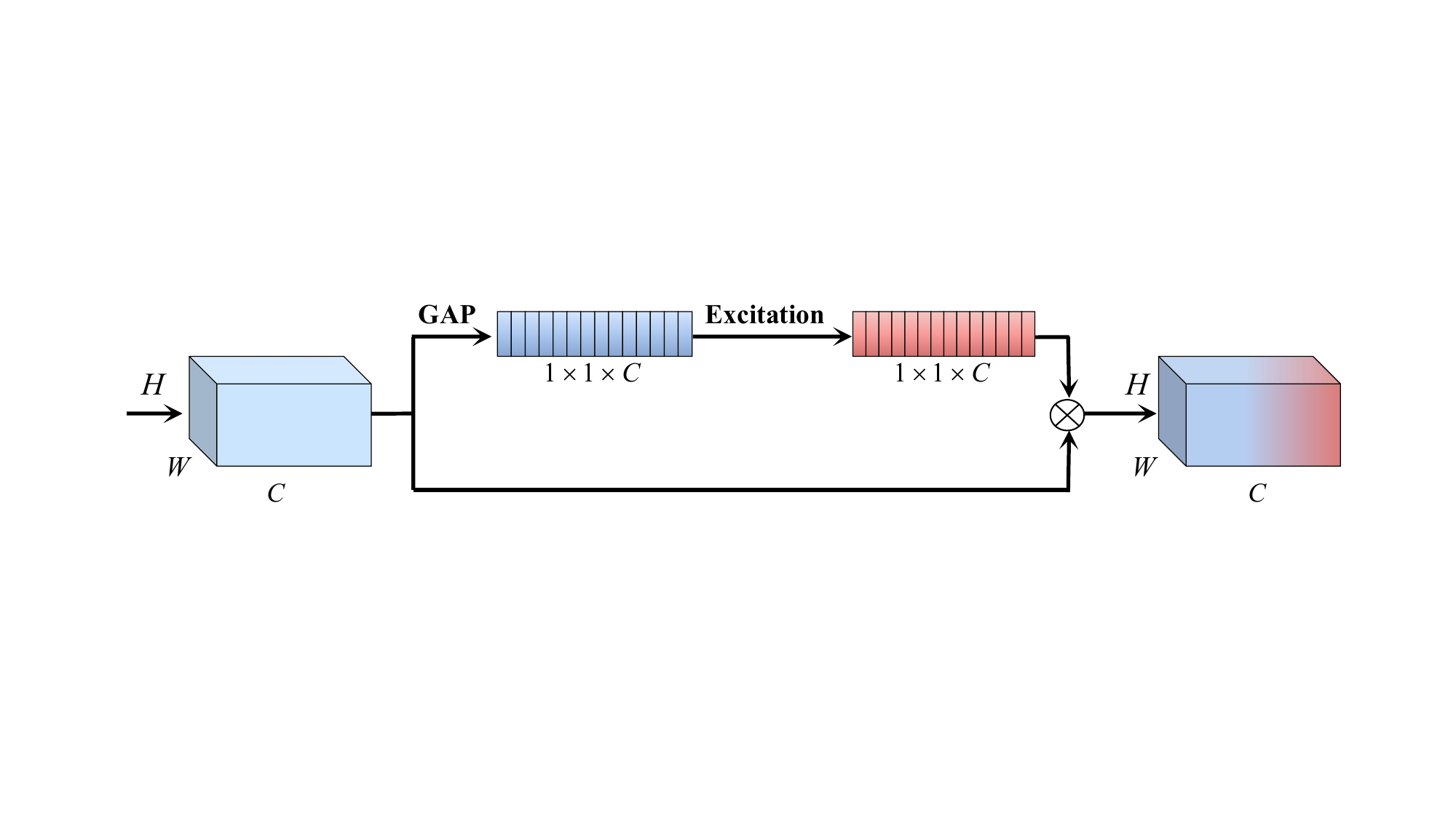}
\end{overpic}
\caption{An illustration of the SE module. }
\label{fig:se_w_module}
\end{figure}
}

\CheckRmv{
\begin{figure}[h]
\centering
\begin{overpic}[width=0.8\linewidth]{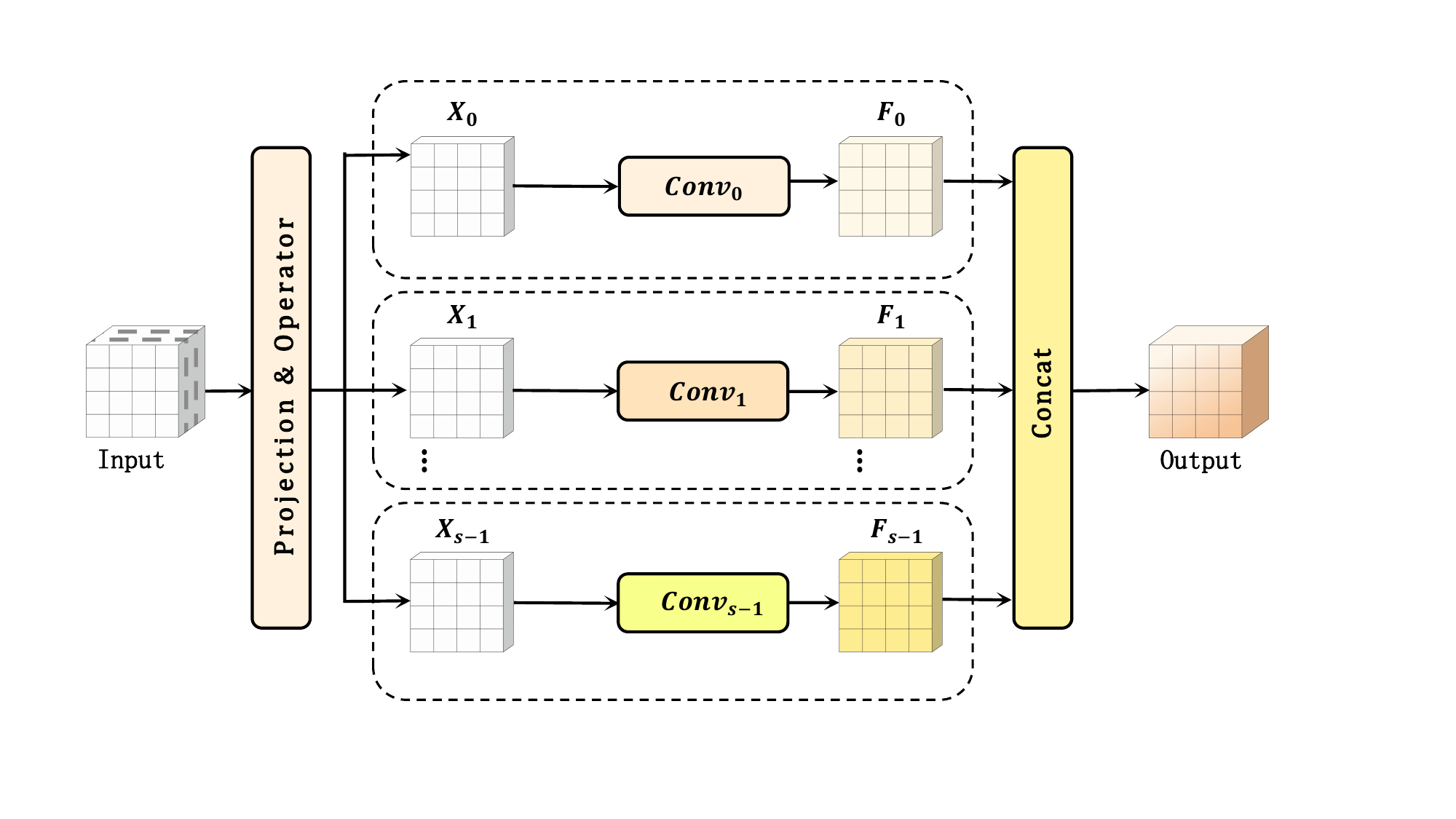}
\end{overpic}
\caption{An illustration of the proposed MBC with $S$ branches, where the concat function is used to concatenate features in the channel dimension. }

\label{fig:MBC_module}
\end{figure}
}
\subsection{Multi-branch and Concat Module}
To fully exploit the multi-scale information from the input feature map effectively in a more flexible way, a novel module namely Multi-branch and Concat (MBC) is proposed in this work. As illustrated by Fig. \ref{fig:MBC_module}, the MBC module can be used as a paradigm of scale transformation and brings new DoF to the attention networks. By allowing the type of projection operators and the number of branches be flexibly adjusted, the MBC module can be implemented in a variety of variants for various application scenarios and tasks. In this work, we mainly focus on two important types of scale transformation operators: Multiplex and Split. Correspondingly, two variations of the MBC module, namely Multiplex and Concat (MUC) and Split and Concat module (SPC), are developed. Both of them can further explore the distribution of multi-scale information on channel dimensions.
The MUC applies a multiplexing operator to the channel dimension of the input tensor, and then divides it into multiple $S$ branches. The input channel dimension is $C$ for each branch, which is the same as the input tensor. By doing this, more abundant positional information from the input tensor is obtained. And the generalization ability of the model can be effectively improved. Next, a multi-scale pyramid convolution structure is used to squeeze the channel dimension of the input tensor and integrate the information at different scales. For the SPC module, however, it applies a splitting operator to reorganize the input feature map more efficiently. Specifically, the input tensor is split into $S$ groups in channel dimension, with $\frac{C}{S}$ channels in each split feature map. Similarly, the multi-scale pyramid convolution structure is used to integrate information at different scales on each channel-wise feature map. By doing this, neighboring scales of context features can be more precisely incorporated and the channel redundancy is reduced by a large margin. In summary, the MUC-based network is more suitable for large backbone CNNs because of its strong generalization ability, and the SPC-based network is more suitable for lightweight networks because it can efficiently reduce the channel redundancy.

\subsubsection{Multiplex and Concat Module}
As illustrated by Fig. \ref{fig:mc_module}, an input feature map $X \in R^{C \times H \times W}$ is multiplexed $S$ times in channel dimension, and then divided into $S$ branches. For each branch, it has $C$ number of input channels as the original input tensor, and the output feature map of each branch has the channel dimension as $C^{'}=\frac{C}{S}$. Note that the number of input channels $C$ should be divisible by $S$. With the MUC design, we can obtain more abundant spatial information from the input tensor. For each branch, it learns the multi-scale spatial information independently and establishes a cross-channel interaction locally. However, an increase in the number of parameters will result from the increase in kernel sizes. In order to process the input tensor at different kernel scales without increasing the computational cost, a method of group convolution is developed and applied to the convolutional kernels in parallel. Furthermore, we design a novel criterion for choosing the group size without increasing the number of parameters. The relationship between the multi-scale kernel and the group size can be written as 
\CheckRmv{
\begin{figure}[t]
\centering
\begin{overpic}[width=0.85\linewidth]{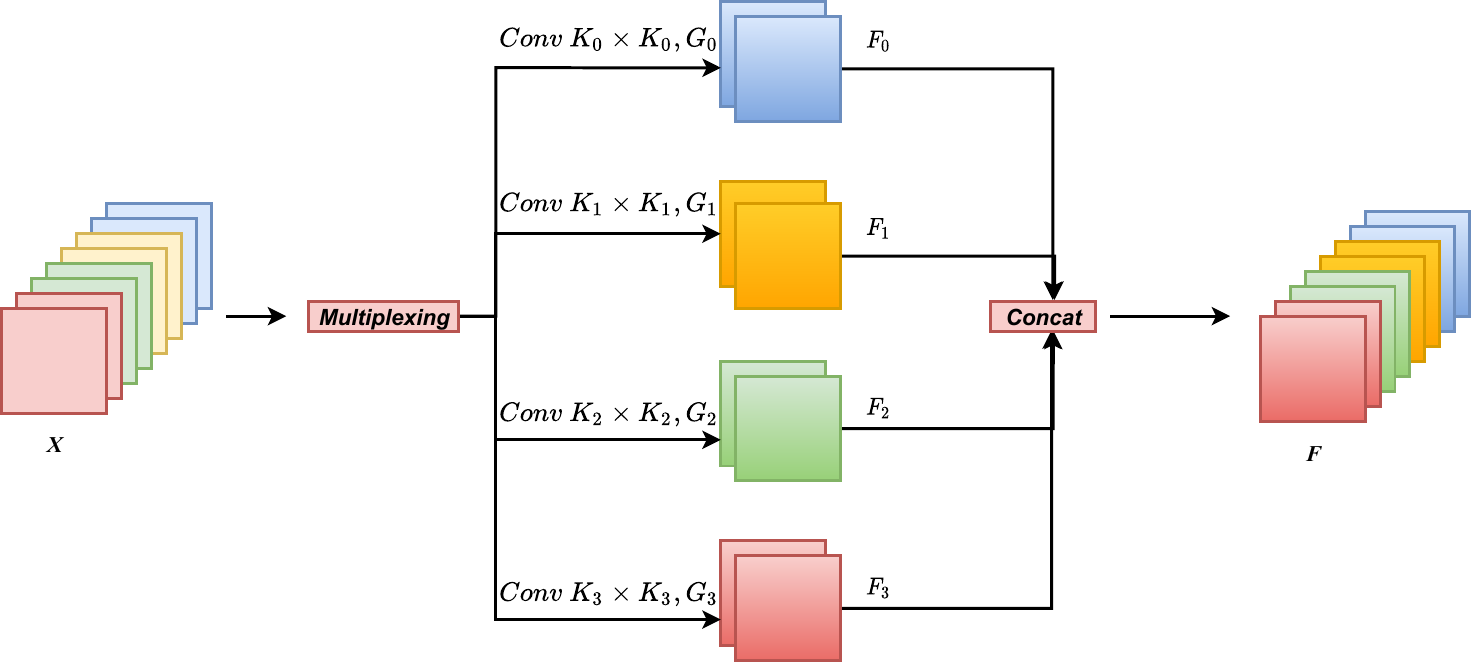}
\end{overpic}
\caption{An illustration of the proposed MUC with the multiplexing rate S sets to 4, where the symbol K is the kernel size and G is the group size. }
\label{fig:mc_module}
\end{figure}
}

\begin{equation}\label{eq:Group_kernel}
G = \left\{
\begin{array}{cl}
2^{\frac{K-1}{2}} & K>3 \\
1 &  K=3 \\
\end{array} \right.
\end{equation}
where the quantity $K$ is the kernel size, and $G$ is the group size. The above equation has been verified by our ablation experiments. Finally, the multi-scale feature map generation function is given by
\begin{equation}
F_{i} = {\rm Conv}(K_{i} \times K_{i}, G_{i})(X) \quad i=0,1,2 \cdots S-1
\end{equation}
where the quantity $K_{i} = 2 \times (i+1) + 1$ is the $i$-th kernel size, $G_{i} = \left\{
\begin{array}{cl}
2^{\frac{K_{i}-1}{2}} & K_{i}>3 \\
1 &  K_{i}=3 \\
\end{array} \right. $ is the $i$-th group size and $F_{i} \in  R^{C^{'} \times H \times W}$ denotes the feature map with different scales. Thus the whole multi-scale preprocessed feature map can be obtained in a concatenating way as
\begin{equation}
F = {\rm Cat}([F_{0}, F_{1}, \cdots, F_{S-1}])	
\end{equation}

\subsubsection{Split and Concat Module}
As shown in Fig. \ref{fig:spc_module}, an input feature map $X \in R^{C \times H \times W}$ is split into $S$ parts, denoted by $[X_{0}, X_{1},\cdots,X_{S-1}]$ along with the channel dimension. For each split part, it has $C^{'} = \frac{C}{S}$ number of channels, and the $i$-th feature map is $X_{i} \in  R^{C^{'} \times H \times W}$ with $i = {0,1,\cdots,S-1}$. It also learns the multi-scale spatial information independently and establishes a cross-channel interaction in a local manner. With this splitting way, we can process the input tensor at multiple scales in parallel. Thus a feature map that contains a single type of kernel can be obtained. The relationship between the group and the multi-scale kernel size is established according to \eqref{eq:Group_kernel}. Finally, the function for multi-scale feature map generation is given by 

\CheckRmv{
\begin{figure}[t]
\centering
\begin{overpic}[width=0.85\linewidth]{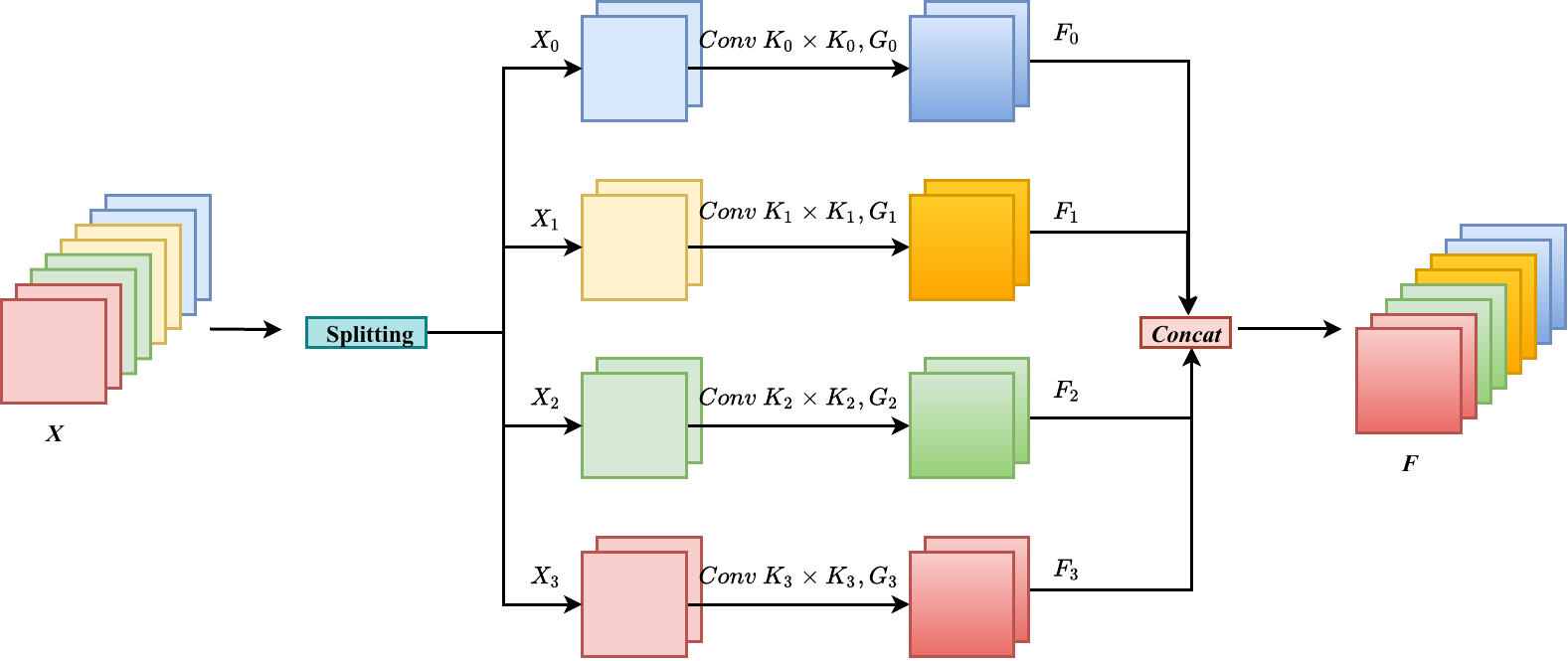}
\end{overpic}
\caption{An illustration of the proposed SPC with the splitting rate sets to 4. }
\label{fig:spc_module}
\end{figure}
}
\begin{equation}
F_{i} = {\rm Conv}(K_{i} \times K_{i}, G_{i})(X_{i}) \quad i=0,1,2 \cdots S-1
\end{equation}
where the quantity $K_{i} = 2 \times (i+1) + 1$ is the $i$-th kernel size, $G_{i} = \left\{
\begin{array}{cl}
2^{\frac{K_{i}-1}{2}} & K_{i}>3 \\
1 &  K_{i}=3 \\
\end{array} \right. $ is the $i$-th group size and $F_{i} \in R^{C^{'} \times H \times W}$ denotes the feature map with different scales. Thus the whole multi-scale preprocessed feature map can be obtained by a concatenating way as
\begin{equation}
F = {\rm Cat}([F_{0}, F_{1}, \cdots, F_{S-1}])	
\end{equation}

\subsection{Multi-branch Attention}
As illustrated by Fig. \ref{fig:mba_module}, a novel multi-branch attention (MBA) module is developed by integrating the MBC and attention module. Correspondingly, the MBA module inherits the advantages of the MBC module and provides a more flexible way of selecting attention modules. Specifically, the MBA module is mainly implemented in four steps. First, the multi-scale feature map on channel-wise is obtained by implementing the proposed MBC module. Second, channel-wise attention vectors are obtained by the attention module for extracting the attention weight of the feature map at different scales. Third, the re-calibrated weight of a multi-scale channel is obtained by using Softmax to re-calibrate the channel-wise attention vector. Fourth, the operation of an element-wise product is applied to the re-calibrated weight and the corresponding feature map. Finally, a refined feature map that is richer in multi-scale feature information is obtained as the output. Therefore, a better interaction between local and global channel attention is achieved by the proposed MBA module.

After processing by the MBC module, the multi-scale feature map $F \in R^{C \times H \times W}$ is obtained. Then, we extract the channel attention weight information from it to generate attention weights with different scales. Mathematically, the attention weight vector is represented as
\begin{equation}
Z_{i} = \Phi (F_{i}), \quad i=0,1,2 \cdots S-1
\end{equation}
where the quantity $Z_{i} \in R^{C^{'} \times 1 \times 1}$ is the attention weight, the symbol $\Phi$ denotes an attention module, which is set as the SE module by default. The attention module is used to obtain the attention weight from the input feature map with different scales. By doing this, the proposed MBA module can fuse context information at different scales and produce better pixel-level attention for high-level feature maps. Further, in order to realize the interaction of attention information and fuse the cross-dimension vectors without destroying the original channel attention vector, the whole multi-scale channel attention vector is obtained in a concatenating way as
\begin{equation}
Z= Z_{0}\oplus Z_{1} \oplus \cdots \oplus Z_{S-1}
\end{equation}
where the symbol $\oplus$ denotes the concat operator, $Z_{i}$ is the attention value from $F_{i}$, and $Z$ is the multi-scale attention weight vector. A soft attention $att_{i}$ as below is used across channels to adaptively select different spatial scales, which is guided by the compact feature descriptor $Z_{i}$.
\begin{equation}
att_{i}= {\rm Softmax}(Z_{i})= \frac{exp(Z_{i})}{\sum_{i=0}^{S-1}exp(Z_{i})}
\end{equation}
\CheckRmv{
\begin{figure}[t]
\centering
\includegraphics[width=\linewidth]{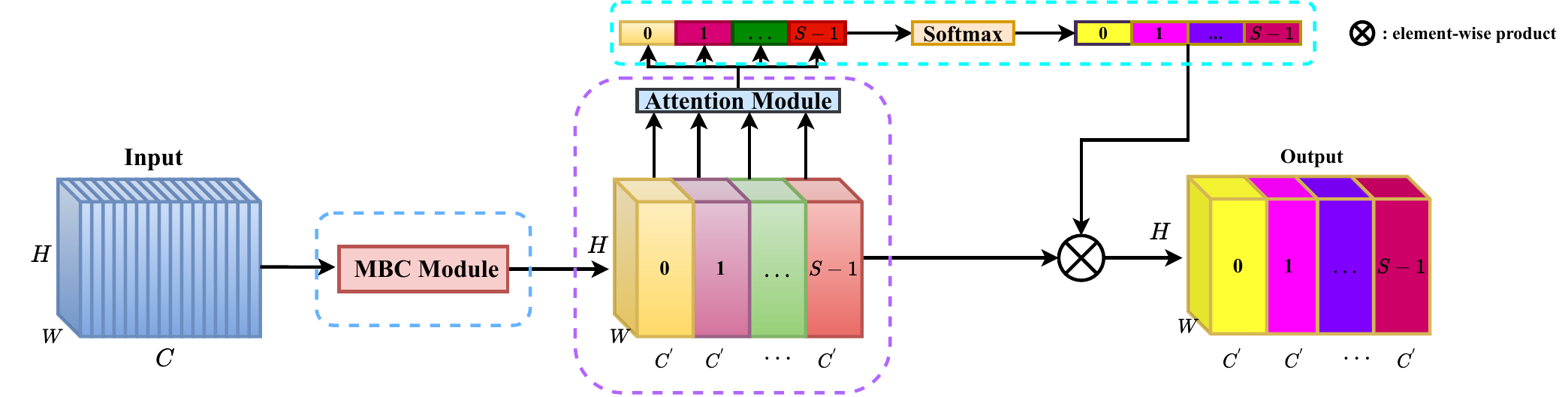}
\caption{The structure of the proposed MBA module, which integrates the MBC and attention modules. }
\label{fig:mba_module}
\end{figure}
}
where the Softmax function is used to obtain the re-calibrated weight $att_{i}$ of the multi-scale channel, which contains the location information and the attention weight in channel. By doing this, the interaction between local and global channel attention is established. Next, the channel attention of feature re-calibration is fused in a concatenation way, and thus the whole channel attention vector can be obtained as
\begin{equation}
att = att_{0} \oplus att_{1} \oplus \cdots \oplus att_{S-1}
\end{equation}
where $att$ represents the multi-scale channel weight after the attention interaction. Then, we multiply the re-calibrated weight of multi-scale channel attention $att_{i}$ with the feature map of the corresponding scale $F_{i}$ as
\begin{equation}
Y_{i} = F_{i} \odot att_{i} \quad i=1, 2,3,\cdots S-1
\end{equation}
where $\odot$ represents the channel-wise multiplication, $Y_{i}$ refers to the feature map that with the obtained multi-scale channel-wise attention weight. The concatenation operator is more effective than the summation because it can integrally maintain the feature representation without destroying the information of the original feature map. Finally, the process to obtain the refined output can be written as
\begin{equation}
Out = {\rm Cat}([Y_{0},Y_{1},\cdots, Y_{S-1}])
\end{equation}

\subsection{Network Design}
As illustrated by Fig.\ref{fig:embanet-block}, a novel block, namely Efficient Multi-branch Attention (EMBA), is developed by substituting the 3x3 convolution in the bottleneck blocks of ResNet with the MBA module at corresponding positions. Then, the multi-scale spatial information and the cross-channel attention are integrated into the EMBA block by the proposed MBA module. Thus, the EMBA block can extract multi-scale spatial information at a more granular level and develop a long-range channel dependency. Next, a novel backbone network EMBANet is further established by stacking the EMBA blocks. The proposed EMBANet inherits the advantages of the EMBA block, and thus it has strong multi-scale representation capabilities and can adaptively recalibrate the cross-dimension channel-wise weight. 
Correspondingly, two versions of the proposed EMBANet, the EMBANet(Small) and EMBANet(Large), are developed accordingly. For the proposed EMBANet(Small), the kernel and group sizes are respectively set as (3, 5, 7, 9) and (1, 4, 8, 16) in the MBC module. On the other hand, the proposed EMBANet(Large) has a higher group size as (32, 32, 32, 32). Because two variants of MBC, i.e. the MUC and SPC are proposed in this work, four networks EMBANet-M(Small), EMBANet-M(Large), EMBANet-S(Small), EMBANet-S(Large) are developed, where the letter 'M' denotes the multiplexing operator and 'S' denotes the splitting operator. That's, the EMBANet-M and EMBANet-S are equipped with the MUC and SPC module, respectively.

\subsection{Discussion}
The proposed MBA module is developed by integrating the MBC and attention modules. The proposed MBA module can be flexibly adjusted according to practical tasks by changing the types of transformation operator and the number of branches. From a technical perspective, the proposed MBA module extends the prior art and serves as a unified framework for various attention methods.
For example, the famous backbone Res2Net in \cite{2019Res2Net} can be interpreted as a special case of the MBA, where its transformation operator can be implemented by connecting the 3x3 convolution in a cascaded way between channels. Similarly, the SKNet \cite{sknet} can be regarded as another instance of the MBA by only setting the number of branches to 2. 
For conventional channel attention modules like ECA or Fca, it can be interpreted as an instance example of the MBA by simply setting the number of branches as 1 and replacing the SE module at corresponding positions. In fact, there are many other models that can be regarded as variants of the MBA. Therefore, the proposed MBA module not only naturally generalizes the pre-processing of scale transformations in the channel-wise dimension but also provides a more flexible design for attention networks. To the best of our knowledge, this is the first attempt to investigate multi-scale feature extraction and enrich the design of the backbone from the DoF perspective in terms of types of transformation operators, attention modules, and the number of branches.

In this paper, we mainly focus on the hand-designed module for CNN architecture. In the future, we will further consider using the method of neural architecture search (NAS) to realize the dynamic balance of MBA sub-modules and find an optimal configuration strategy on different sub-tasks.

\section{Experiments}\label{sec:experiments}
\subsection{Implementation Details}
\label{experiments_details}
In this work, the Pytorch framework is employed to implement experiments. 
For image classification tasks, the ResNet \cite{he2016deep} is used as the backbone and ImageNet as the dataset\cite{krizhevsky2012imagenet}. Similar to the prior art, the training configuration is set according to the reference in \cite{2019Res2Net,hu2018senet,2020FcaNet}, the standard data augmentation scheme is implemented, and the size of the input tensor is cropped to 224 $\times$ 224 by randomly horizontal flipping and normalization. The optimization was performed by using stochastic gradient descent (SGD) with a weight decay of 1e-4, momentum of 0.9, and a mini-batch size of 256. The label-smoothing regularization \cite{labelsmooth_inceptionv2} is used with the coefficient value of 0.1 during training. The learning rate was initially set as 0.1 and decreased by a factor of 10 after every 30 epochs for 100 in total.
For object detection tasks, the default setting is that the shorter side of the input image is resized to 800, the SGD is used with a weight decay of 1e-4, momentum of 0.9, and the batch size is 2 per GPU within 12 epochs. The learning rate is 0.01 and decreased by a factor of 10 at the 8th and 11th epochs, respectively. 
For instance segmentation tasks, the settings of the training configuration and dataset are similar to those are used for object detection. Finally, all detectors are implemented by the MMDetection toolkit, and all models are trained on 8 Titan RTX GPUs.
\begin{table}[h]
	\setlength{\tabcolsep}{5mm}
	\centering
	\caption{The performance of the DoF for the proposed network on ImageNet.}
	\resizebox{0.85\textwidth}{!}{
		\begin{tabular}{l|l|c|c|c|c|c} \toprule
			Network     &Backbone   &Attention Module &Parameter  & FLOPs & Top-1 Acc. ($\%$)&Top-5 Acc. ($\%$)\\ \midrule
			SENet\cite{hu2018senet} &\multirow{2}{18mm}{ResNet-50} & SE & 28.07M  & 4.13G    & 76.71          & 93.38             \\ 
			EMBANet-M(Large) & & SE & 27.90M  & 4.72G & 78.65 & 94.25  \\ 
			\midrule
			
			ECANet\cite{20ecanet}  &\multirow{2}{18mm}{ResNet-50} & ECA & 25.56M  &4.13G   & 77.48   & 93.68       \\ 
			EMBANet-M(Large)V2      & & ECA & 27.86M  & 4.72G & 78.81 & 94.34  \\
			\midrule
			
			FcaNet\cite{2020FcaNet} &\multirow{2}{18mm}{ResNet-50}  &Fca  & 28.07M  &4.13G  & 78.52   & 94.14 \\
			EMBANet-M(Large)V3  & &Fca   &27.90M  &4.72G &78.97 & 94.24  \\
			\midrule
			
			SENet\cite{hu2018senet} &\multirow{2}{18mm}{ResNet-101} & SE & 49.29M &7.86G &77.62 &93.93    \\ 
			EMBANet-M(Large) & & SE &49.59M &8.97G &79.38 &94.58   \\ 
			\midrule
			
			ECANet\cite{20ecanet}  &\multirow{2}{18mm}{ResNet-101} & ECA & 44.55M &7.86G &78.65 &94.34     \\ 
			EMBANet-M(Large)V2      & & ECA &49.51M &8.97G &79.71 &94.67  \\
			\midrule
			
			FcaNet\cite{2020FcaNet} &\multirow{2}{18mm}{ResNet-101}  &Fca  & 49.29M & 7.86G & 79.64 & 94.63 \\
			
			EMBANet-M(Large)V3  & & Fca   &49.59M &8.97G &79.87 &94.68 \\
			
			\bottomrule
	\end{tabular}}
	\label{table1:image-dof}
\end{table}
\subsection{The Degrees of Freedom}
To demonstrate the benefits of the DoF from the proposed MBA-based framework, we performed a series of dynamic transformations on the ImageNet dataset. 
Two widely used CNNs, ResNet-50\cite{he2016deep} and ResNet-101 \cite{he2016deep}, are employed as backbone models. 
The proposed MBA framework can be flexibly equipped with various attention modules such as SE, ECA, and Fca. 
In particular, the EMBANet-M(Large)V2 and EMBANet-M(Large)V3 are developed by employing the ECA and Fca respectively. As illustrated in Table \ref{table1:image-dof}, the proposed MBA-based networks outperform their counterparts in terms of Top-1 accuracy for the ResNet-50 backbone based models by 1.94$\%$, 1.33$\%$, and 0.45$\%$ respectively. For the ResNet-101 based models, the proposed MBA-based networks can outperform their counterparts by 1.76$\%$, 1.06$\%$, and 0.23$\%$ in terms of Top-1 accuracy, respectively. It should be noted that the proposed EMBANet-M(Large)V3 outperforms the SENet in terms of Top-1 accuracy by more than 2$\%$ on both the ResNet-50 and ResNet-101 based backbones. The results demonstrated the overall flexibility and effectiveness of the proposed MBA module. Although the experiments are based on the EMBANet-M(Large), similar results can be obtained for other proposed networks.
\begin{table}[h]
	\setlength{\tabcolsep}{5mm}
	\centering
	\caption{Performance comparisons with other CNNs on ImageNet. }
	\resizebox{0.85\textwidth}{!}{
		\begin{tabular}{l|l|c|c|c|c} \toprule
			Network         & Backbone  &Parameter  & FLOPs & Top-1 Acc. ($\%$)&Top-5 Acc. ($\%$)\\ \midrule
			ResNet\cite{he2016deep}          & \multirow{11}{18mm}{ResNet-50}  & {25.56M} &4.12G & 76.15  & 92.87    \\ 
			Res2Net-50\cite{2019Res2Net}  & &25.70M   &4.28G     & 77.99          &  93.85           \\ 
			SENet\cite{hu2018senet}    & & 28.07M  & 4.13G    & 76.71          & 93.38             \\ 
			Res2Net-50+SE\cite{2019Res2Net}   		& & 28.21M   & 4.29G    &  78.44       & 94.06            \\ 
			CBAM\cite{cbam}           & & 28.07M  &4.14G    & 77.34          & 93.69             \\ 
			A$^{2}$-Net\cite{doubleAAnet}      & & 33.00M  &6.50G    & 77.00          & 93.50             \\ 
			GCNet\cite{2019GCNet}          & & 28.11M  &4.13G    & 77.70          & 93.66               \\  
			AANet\cite{AAnet}  & & 25.80M  &4.15G    & 77.70          & 93.80               \\ 
			ECANet\cite{20ecanet}        & & 25.56M  &4.13G    & 77.48          & 93.68       \\ 
			SK-ResNet50\cite{sknet}	& & 26.15M	&4.19G	&77.55	& 93.82 \\
			FcaNet\cite{2020FcaNet}        &  & 28.07M  &4.13G  & 78.52   & 94.14 \\ \midrule
			EMBANet-S(Small)  &\multirow{4}{18mm}{ResNet-50} & \textbf{16.33M}  & \textbf{2.60G} & 77.33 & 93.54 \\ 
			EMBANet-M(Small)     & & 22.56M  & 3.62G   & 77.55  & 93.65     \\
			EMBANet-S(Large)  & & 25.27M & 4.29G & 78.35 & 94.07 \\
			EMBANet-M(Large)  & & 27.90M  & 4.72G & 78.65 & 94.25  \\ \midrule
			EMBANet-M(Large)V2  &\multirow{2}{18mm}{ResNet-50} & 27.86M  & 4.72G & 78.81 & 94.34  \\
			EMBANet-M(Large)V3  & & 27.90M  & 4.72G & \textbf{78.97} & 94.24  \\
			\bottomrule
	\end{tabular}}
	\label{table2:image-prec-sota}
\end{table}
\subsection{ImageNet}\label{sec:Experiments_ImageNet}
In this section, the performance of the proposed MBA based networks is compared with several \sArt attention methods, such as the SENet \cite{hu2018senet}, CBAM \cite{cbam}, $A^{2}$-Net \cite{doubleAAnet}, GCNet \cite{2019GCNet}, ECANet \cite{20ecanet}, FcaNet \cite{2020FcaNet}, and SKNet \cite{sknet} on ImageNet dataset~\cite{russakovsky2015imagenet}. The ResNet-50 is employed as the backbone network in these experiments.

As shown by \tabref{table2:image-prec-sota}, the proposed EMBANet-M(Large) outperforms the SENet, ECANet, and FcaNet in terms of Top-1 accuracy by approximately 1.94$\%$, 1.17$\%$ and 0.13$\%$ respectively, while maintaining almost the same computational cost. The proposed EMBANet-M(Large)V3 improves the Res2Net and SKNet in terms of Top-1 accuracy by about 0.98$\%$ and 1.42$\%$ respectively. Meanwhile, with almost the same complexity, the proposed EMBANet-M(Large)V3 outperforms the Res2Net+SE and the FcaNet by 0.53$\%$ and 0.45$\%$ in terms of Top-1 accuracy respectively. The proposed EMBANet-S(Small) outperforms the SENet-50 by 0.62$\%$, while requiring 41.8$\%$ fewer parameters and a 41.4$\%$ lower computational cost. Furthermore, with nearly the same complexity, the proposed EMBANet-S(Large) outperforms the ECANet by 0.87$\%$ in terms of Top-1 accuracy. 
In summary, the proposed EMBANet-S type networks can reduce channel redundancy and combine channels effectively while still maintaining the quality of multi-scale feature representations, which is more suitable for a lightweight network structure. And the proposed EMBANet-M type networks can provide considerable performance gains for various vision tasks.
\CheckRmv{
\begin{table}[h]
\setlength{\tabcolsep}{5mm}
\centering
\caption{Top-1 accuracy on CIFAR-100. } 
\resizebox{0.85\textwidth}{!}{
		\begin{tabular}{l|c|c} \toprule
			Backbone       & Parameter & Top-1 acc($\%$) \\ \midrule
			ResNet-50     &23.71M & 77.67 \\ 
			SENet-50     &26.22M  & 77.91 \\
			EMBANet-S(Small)-50  &\textbf{14.48M} & 79.02 \\ 
			EMBANet-M(Small)-50  &20.71M & \textbf{79.83} \\    
			\bottomrule
	\end{tabular}}
	\label{tab:cifar}
\end{table}
}
\subsection{CIFAR}
\label{sec:Experiments_CIFAR}
We conduct experiments on the CIFAR-100 dataset~\cite{krizhevsky2009learning}, which contains 50k training images and 10k testing images from 100 classes. We employ the ResNet-50 as the backbone, and only replace the original basic blocks with our proposed MBA module while keeping the other configurations unchanged. As shown by \tabref{tab:cifar}, the proposed EMBANet-S(Small) achieves a large margin of 1.35$\%$  and 1.11$\%$ higher Top-1 accuracy over ResNet-50 and SENet-50, and reduces the number of parameters by 38.9$\%$ and 44.8$\%$ respectively. The proposed EMBANet-M(Small) further improves Top-1 accuracy by 1.92$\%$ and 2.16$\%$ as compared to the SENet and the ResNet respectively, while using much fewer parameters. In summary, efficient and effective performance can be achieved simultaneously by both the proposed EMBANet-M and EMBANet-S types of networks.

\begin{table}[h]
\centering
\setlength{\tabcolsep}{5mm}
\caption{Comparison of deeper and wider CNNs on ImageNet in terms of Top-1 and Top-5 accuracy. 
}
	\resizebox{0.8\textwidth}{!}{
		\begin{tabular}{lcc} \toprule
			Network        & Top-1 Acc. ($\%$)&Top-5 Acc. ($\%$)\\ \midrule
			ResNet-200\cite{he2016deep}   & 78.20 & 94.00 \\ 
			Res2Net-101\cite{2019Res2Net}    & 79.19 & 94.43\\ 
			Inception-v3\cite{szegedy2016rethinking}   & 77.45 & 93.56 \\ 
			ResNeXt-101\cite{xie2017aggregated}   & 78.80 & 94.40 \\ 
			SK-ResNet101\cite{sknet} &78.84 &94.29 \\
			DenseNet-264 (k=32)\cite{huang2017densely}  & 77.85 & 93.78 \\ 
			DenseNet-161 (k=48)\cite{huang2017densely} & 77.65 & 93.80 \\ \hline
			EMBANet-M(Large)    &79.38 &94.58  \\
			EMBANet-M(Large)V2  &79.71 &94.67 \\ 
			EMBANet-M(Large)V3  &\textbf{79.87} &\textbf{94.68} \\
			\bottomrule
	\end{tabular}}
	\label{tab:deep_wider_imagenet}
\end{table}
\section{Ablation Analysis}
\subsection{Going Deeper and Wider with \ourM}
Deeper and wider networks have demonstrated a stronger representation capability~\cite{he2016deep,xie2017aggregated,szegedy2017inception} for vision tasks. As presented in Table \ref{tab:deep_wider_imagenet}, we mainly compare the performance of the proposed three variants that are based on the EMBANet-M architecture with several classical and more complicated CNNs, such as the Res2Net \cite{2019Res2Net}, Inception-v3~\cite{szegedy2016rethinking}, ResNeXt~\cite{xie2017aggregated}, SKNet \cite{sknet}, and DenseNet \cite{huang2017densely}. The proposed three variants employ ResNet-101 as the backbone network. 
To implement a more fair comparison, we duplicated the SKNet-101 with the same configuration and dataset. 
As shown in Table \ref{tab:deep_wider_imagenet}, the proposed EMBANet-M(Large) outperforms the DenseNet-264 (k=32), DenseNet-161 (k=48) and Inception-v3 in terms of Top-1 accuracy, respectively, by about 1.53$\%$, 1.73$\%$ and 1.93$\%$. What's more, the proposed EMBANet-M(Large)V2 achieves 1.51$\%$ and 0.91$\%$ gains in terms of Top-1 accuracy as compared to ResNet-200 (the one with deeper layers) and ResNeXt-101 (the one employing more convolutional filters and expensive group convolutions), respectively. The proposed EMBANet-M(Large)V3 outperforms more competitive models such as Res2Net-101 and SKNet-101 in terms of Top-1 accuracy by 0.68$\%$ and 1.03$\%$, respectively. The above results demonstrate that the proposed EMBANet-M type networks can effectively strengthen multi-scale feature extraction and compact the internal representations in the channel dimension. Meanwhile, the proposed EMBANet-M type networks can perform favourably against state-of-the-art CNNs which are even deeper and wider in network architecture. These results also verified that the proposed MBA module is very flexible and can be integrated with deeper models when necessary to achieve better performance. 

\CheckRmv{
	\begin{figure}[h]
		\centering
		\begin{overpic}[width=0.75\linewidth]{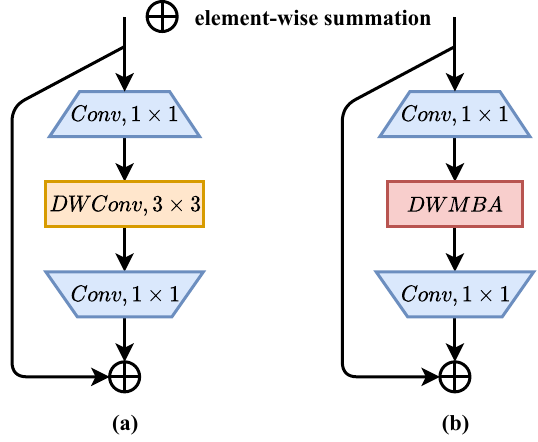}
		\end{overpic}
		\caption{
			(a) Inverted residual block proposed in MobileNetV2. 
			(b) our novel lightweight block from which the proposed EMBANet-L network is developed. }
		\label{fig:nbv2_module}
	\end{figure}
}
\begin{table}[h]
	\centering
	
	\caption{Comparison of classical lightweight methods on ImageNet in terms of network parameters and Top-1 accuracy. }
	\setlength{\tabcolsep}{5mm}{
		\resizebox{0.8\textwidth}{!}{
			\begin{tabular}{lcc} \toprule
				Backbone            & Parameters   & Top-1 Acc.($\%$)  \\ \midrule
				MobileNetV2\cite{mbv2}  & 3.50M & 72.00    \\
				MobileNetV2 + SE & 3.89M & 72.42 \\
				EMBANet-L  & 3.75M     & \textbf{73.40} \\
				\bottomrule
	\end{tabular}}}
	\label{tab:mbv2}
\end{table}

\subsection{Going Lightweight with EMBANet}
\label{sec:Experiments_mbv2}
As illustrated by Fig. \ref{fig:nbv2_module}, we create a new lightweight module called depthwise-MBA (DWMBA) by substituting the 3x3 deep separable convolution at corresponding positions in MobileNetV2\cite{mbv2} with the proposed MBA. In particular, the proposed DWMBA module consists of the proposed SPC module and the SE attention. The proposed SPC module turns each convolution with different scales into a pyramid structure, such as 3x3, 5x5, 7x7, and 9x9 convolutions, into a depth-wise separable convolution. Correspondingly, a novel lightweight network named EMBANet-L is further established by stacking the lightweight blocks in the MobileNetV2 style. During the training process, we use the standard SGD optimizer with the decay and momentum parameter set as 0.9 for all models. The weight decay rate is set to $4 \times 10^{-5}$. A cosine learning schedule with an initial learning rate of 0.05 is adopted, and the batch size is 256. We use MobileNetV2 \cite{mbv2} as our backbone, and all models are trained with 200 epochs. For data augmentation, we use the same methods as in MobileNetV2. As shown in Table \ref{tab:mbv2}, our proposed EMBANet-L improves Top-1 accuracy by approximately 1.4$\%$ and 0.98$\%$ over the original MobileNetV2 and the updated MobileNetV2 with SENet, respectively. Based on the above results, the efficiency and effectiveness of the proposed MBA module for lightweight CNN architectures is verified.

\begin{table}[h]
	\centering
	\caption{Acc. with the change of re-calibrating functions. }
	\setlength{\tabcolsep}{5mm}{
		\resizebox{0.8\textwidth}{!}{
			\begin{tabular}{ccc} \toprule
				Model  & Excitation      & Top-1 Acc($\%$)  \\ 
				\midrule
				\multirow{2}{30mm}{EMBANet-M(Small)}  & Sigmoid  & 79.57     \\ 
				&      Softmax   &  \textbf{79.83}     \\ \hline
				\multirow{2}{30mm}{EMBANet-S(Small)} & Sigmoid  & 78.92     \\ 
				&      Softmax   &  \textbf{79.02}     \\ \bottomrule
	\end{tabular}}}
	\label{ablation_act}
\end{table}	
\begin{table}[h]
	\centering
	\caption{Accuracy of the SPC and the MUC module on CIFAR-100. }
	\setlength{\tabcolsep}{5mm}{
		\resizebox{0.8\textwidth}{!}{
			\begin{tabular}{lccc} \toprule
				Model  & Module   & Parameter & Top-1 Acc($\%$)   \\
				\midrule
				\multirow{3}{30mm}{EMBANet-S(Small)}
				& SPC   & 14.47M  & 78.86  \\
				& SE  & 23.87M   & 78.69 \\ 
				& SPC+SE & 14.48M   & 79.02   \\ \midrule
				\multirow{3}{30mm}{EMBANet-M(Small)} 
				& MUC   &20.70M  &78.81   \\
				& SE   &23.87M  &78.86 \\ 
				& MUC+SE  &20.71M  &\textbf{79.83} \\
				\bottomrule
	\end{tabular}}}
	\label{ablation_mbc_module}
\end{table}

\subsection{The Impacts of the Re-calibrating Function} 
In order to re-calibrate the channel-wise attention vector more effectively, we assessed the choice of the excitation function for the re-weight mechanism. We mainly consider two commonly used options: Softmax and Sigmoid. For the SPC and MUC modules, we compared the performance of using Softmax with that of using the Sigmoid operation on the CIFAR-100 dataset. As shown in Table \ref{ablation_act}, the performance of the Softmax is better than that of the Sigmoid.

\subsection{The Impacts of the SPC and the MUC module}
As shown by Fig \ref{fig:mba_module}, the proposed EMBANet consists of the MBC and the SE attention modules. In order to explicitly verify that benefits resulted from the MBC module, we conducted this experiment and employed the EMBANet-S(Small) and the EMBANet-M(Small) as baselines on the CIFAR-100 dataset. As illustrated by Table \ref{ablation_mbc_module}, the word 'SPC' means to remove the SE attention module and only keeping the SPC, the word 'MUC' means to remove the SE attention module and only keeping the MUC, the word 'SE' means that the number of branches in the MBC module is set to 1, which can be interpreted as removing the MBC module. Besides, the word 'SPC+SE' means that both the SPC and SE attention modules are equipped, and the word 'MUC+SE' means that both the MUC and SE attention modules are equipped. The experimental results verified that the proposed MBC is more effective than the SE module in terms of performance improvement. Thus, the proposed MBC-based framework brings flexibility and effectiveness to attention network design.
\begin{table}[h]
	\centering
	\caption{Performance comparisons of object detection on COCO val2017. }
	\setlength{\tabcolsep}{5mm}
	\resizebox{0.85\textwidth}{!}{
		\begin{tabular}{llcccccccc} \toprule
			Backbone & Detectors  &Parameter &FLOPs & AP & $AP_{50}$ & $AP_{75}$ & $AP_{S}$ & $AP_{M}$ & $AP_{L}$ \\ \midrule
			ResNet-50\cite{he2016deep}  & \multirow{5}{10mm}{Faster-RCNN}  & 41.53M & 207.07G & 36.4 & 58.2 & 39.5 &21.8 & 40.0 & 46.2   \\  
			SENet-50\cite{hu2018senet}          & & 44.02M & 207.18G & 37.7 & 60.1 & 40.9 & 22.9 & 41.9 & 48.2 \\
			ECANet-50\cite{20ecanet}   & & 41.53M & 207.18G & 38.0 & 60.6 & 40.9 & 23.4 & 42.1 & 48.0 \\   
			FcaNet-50\cite{2020FcaNet}  & & 44.02M & 215.63G & 39.0 & 61.1 & 42.3 & 23.7 &42.8 &49.6 \\
			EMBANet-M(Large)-50  &  &43.85M  &219.64G   &\textbf{40.9}  &\textbf{62.1} &\textbf{44.6}  &23.6  &\textbf{44.5} & \textbf{54.0}   \\
			
			\hline  
			ResNet-50\cite{he2016deep}  & \multirow{6}{10mm}{Mask-RCNN}  & 44.18M & 275.58G  & 37.2  & 58.9  & 40.3 & 22.2  & 40.7  & 48.0   \\
			SENet-50\cite{hu2018senet}  &     & 46.67M  & 275.69G  & 38.7   & 60.9  & 42.1   & 23.4   & 42.7   & 50.0   \\
			Non-local\cite{Non-local_long}  &      & 46.50M  & 288.70G & 38.0   & 59.8  & 41.0   & -     & -      & -    \\
			GCNet-50\cite{2019GCNet}                   &      & 46.90M & 279.60G  & 39.4   & 61.6  & 42.4   & -     & -      & -    \\
			ECANet-50\cite{20ecanet}      &      & 44.18M  & 275.69G & 39.0   & 61.3  & 42.1   & 24.2  & 42.8   & 49.9  \\
			FcaNet-50\cite{2020FcaNet} & &46.66M &261.93G &40.3 & 62.0 & 44.1 &25.2 & 43.9 & 52.0 \\ 
			EMBANet-M(Large)-50  & & 46.50M & 271.10G  &\textbf{41.4}  &\textbf{62.3}  &\textbf{45.3} &23.6  &\textbf{45.1}  &\textbf{54.6}  \\	
			\hline
			ResNet-50\cite{he2016deep}  & \multirow{3}{10mm}{RetinaNet}  & 37.74M & 239.32G & 35.6  & 55.5  & 38.2   & 20.0   & 39.6   & 46.8      \\
			SENet-50\cite{hu2018senet}   & & 40.25M & 239.43G & 37.1 & 57.2 & 39.9 & 21.2 & 40.7 & 49.3  \\
			EMBANet-M(Large)-50  &  &40.07M  &251.89G  &\textbf{39.6} &\textbf{59.4} &\textbf{42.3}  &21.2  &\textbf{43.4}  &\textbf{52.9}  \\
			
			\bottomrule
	\end{tabular}}
	\label{detection_result}
\end{table}

\newcommand{\RowsT}[1]{{\multirow{3}{*}{\begin{tabular}[c]{@{}c@{}}#1\\ ($\%$)\end{tabular}}}}
\begin{table}[h]
	\centering
	\setlength{\tabcolsep}{5mm}
	\caption{Average precision of object detection on COCO. }
	\resizebox{0.8\textwidth}{!}{
		\begin{tabular}{lccccc} \toprule
			&            & Small & Medium & Large & All  \\   \midrule
			SENet-50 & \RowsT{AP}   & 23.4   & 42.7   & 50.0 & 38.7   \\
			EMBANet-S(Large)-50  &  & 23.8   & 44.7  & 54.3  & 41.1  \\ 
			Improve.  &            & \textbf{0.4} & \textbf{2.0}  & \textbf{4.3}  & \textbf{2.4}   \\ 
			\hline
			SENet-50 & \RowsT{AP}   & 23.4   & 42.7   & 50.0 & 38.7   \\
			EMBANet-M(Large)-50  &  & 23.6   & 45.1  & 54.6  & 41.4  \\ 
			Improve.  &            & \textbf{0.2} & \textbf{2.4}  & \textbf{4.6}  & \textbf{2.7}   
			\\ 
			\bottomrule
	\end{tabular}}
	\label{tab:object_detection_size}
\end{table}

	\begin{figure}[h]
		\centering
		\begin{overpic}[width=0.6\linewidth]{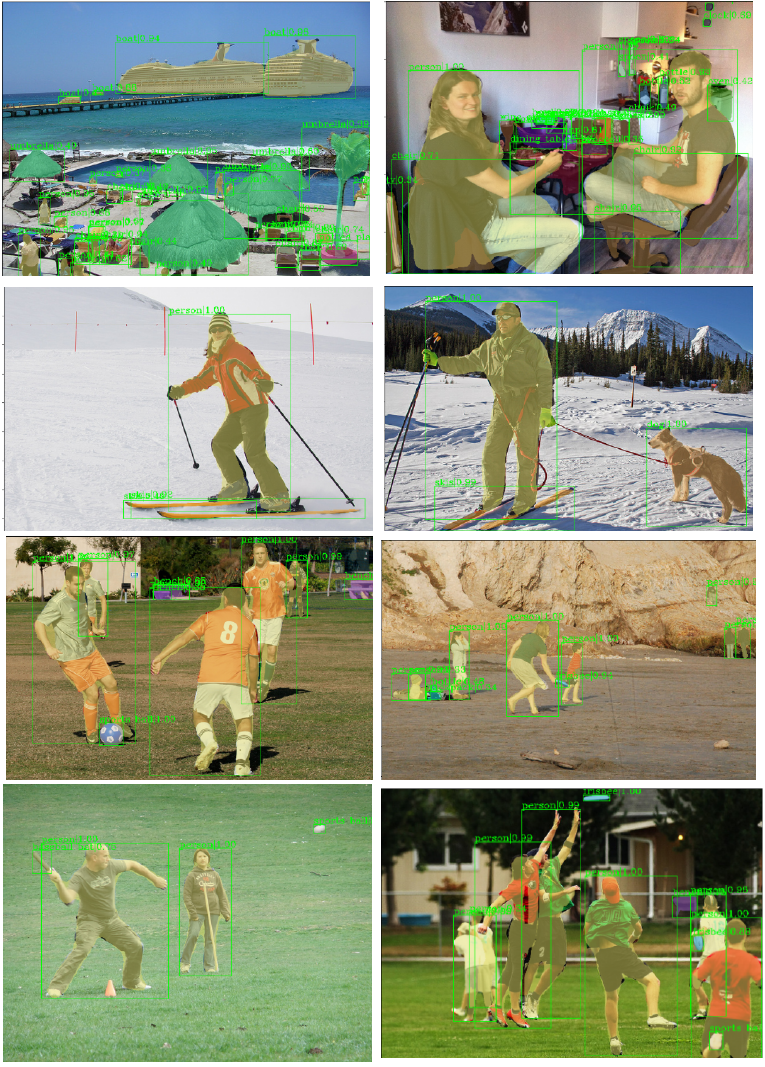}
		\end{overpic}
		\caption{Visualization results of object detection and instance segmentation for EMBANet-S(Large)-50 on MS-COCO val set. }
		\label{fig:coco_embanet_s}
	\end{figure}

	\begin{figure}[h]
		\centering
		\begin{overpic}[width=0.6\linewidth]{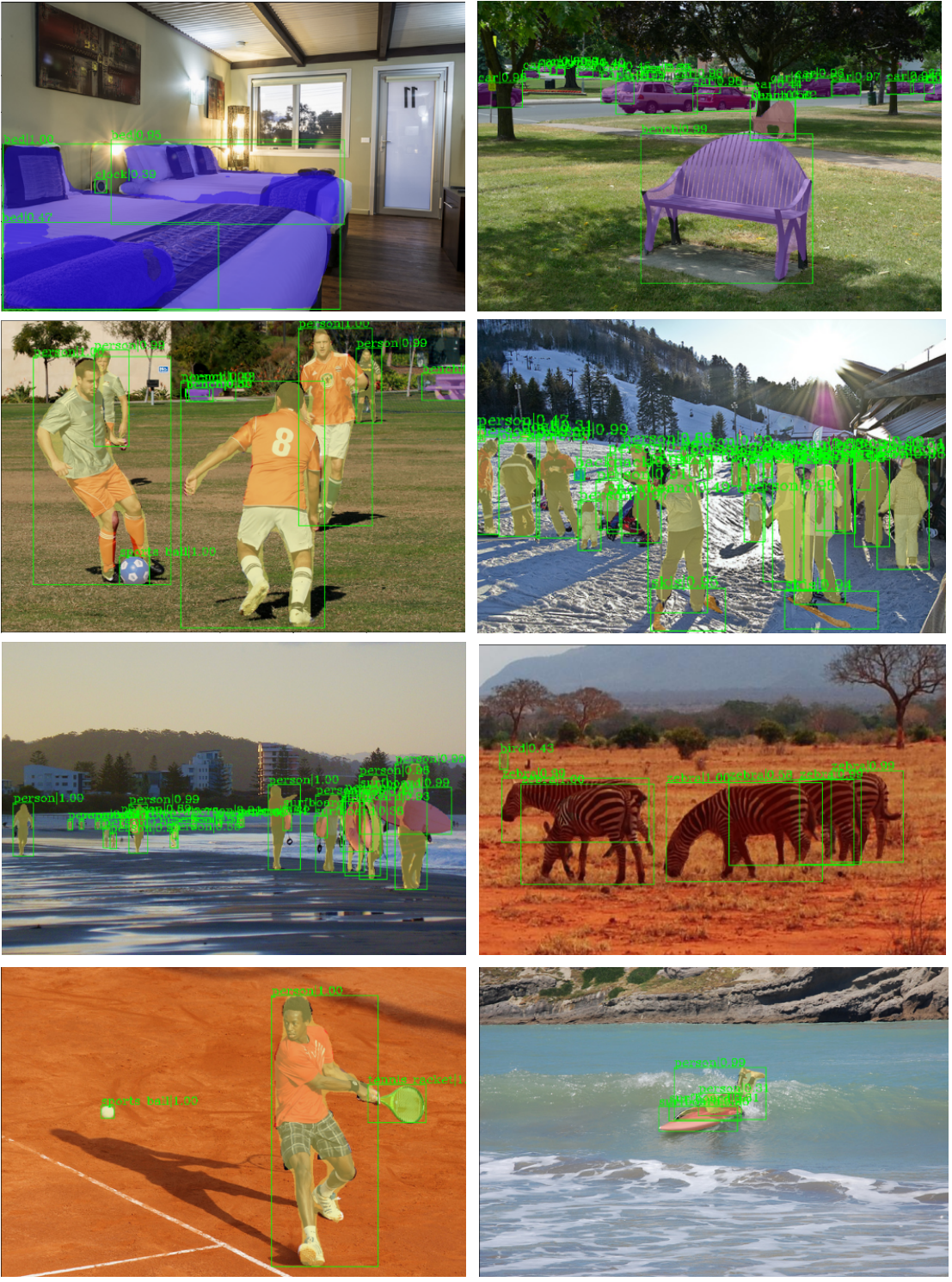}
		\end{overpic}
		\caption{Visualization results of object detection and instance segmentation for EMBANet-M(Large)-50 on MS-COCO val set. }
		\label{fig:coco_embanet_m}
	\end{figure}
\begin{table}[htp]
	\setlength{\tabcolsep}{5mm}
	\centering
	\caption{Instance segmentation performances on COCO. }
	\resizebox{0.8\textwidth}{!}{
		\begin{tabular}{lccc} \toprule
			Backbone           & $AP$ & $AP_{50}$ & $AP_{75}$  \\ \midrule
			ResNet-50 & 34.1 & 55.5 & 36.2 \\ 
			SENet-50 & 35.4  & 57.4  & 37.8   \\
			ResNet-50 + 1 NL-block\cite{Non-local_long} & 34.7 & 56.7 & 36.6 \\ 
			GCNet & 35.7 & 58.4 & 37.6 \\ 
			ECANet & 35.6  & 58.1 & 37.7 \\
			FcaNet & 36.2 & 58.6 & 38.1 \\ \hline 
			EMBANet-S(Large) & 36.7 &58.8 & 39.0  \\
			EMBANet-M(Large) & \textbf{37.1}  &\textbf{59.0}  & \textbf{39.5} \\ 
			\bottomrule
	\end{tabular}}
	\label{tab:instance_segmantation}
\end{table}

\section{Object Detection}\label{sec:object_det}
In order to demonstrate the generalization ability of our proposed \ourM~on downstream vision tasks, experiments on object detection are implemented. The ResNet-50 along with FPN\cite{FPN} is used as the backbone model. The three representative detectors, Faster RCNN~\cite{ren2015faster}, Mask RCNN~\cite{he2017mask}, and RetinaNet~\cite{retinanet} are used on the MS-COCO \cite{lin2014microsoft} dataset. As shown by \tabref{detection_result}, the EMBANet-M(Large)-50 achieves the best performance compared with the other attention methods. 
When compared to the ResNet-50, the proposed EMBANet-M(Large)-50 improves the mean average precision (AP) by around 4$\%$ on the three detectors mentioned above. 
As compared to the FcaNet, the AP can be improved by 1.9$\%$ and 1.1$\%$ by our proposed EMBANet-M(Large)-50 on both Faster-RCNN and Mask-RCNN detector with almost the same computational complexity. It is worth noting that the most compelling performance improvement appears in the measurement of $AP_L$. The results demonstrated that the proposed EMBANet-M(Large)-50 has good generalization ability and can be easily applied to other downstream tasks. To evaluate the performance of the EMBANet-S architecture, we employ the EMBANet-S(Large) as the backbone and the Mask-RCNN as the detector. As shown by \tabref{tab:object_detection_size}, the proposed EMBANet-S(Large)-50 achieves 0.4$\%$, 2.0$\%$ and 4.3$\%$ gains over the SENet-50 respectively on the small, medium, and large objects. And the proposed EMBANet-M(Large)-50 achieves 0.2$\%$, 2.4$\%$, and 4.6$\%$ gains respectively on small, medium, and large objects. The results verified that both the proposed EMBANet-S and EMBANet-M architectures can be effectively migrated to various vision tasks and improved performance can be achieved.
	

\section{Instance Segmentation}
Instance segmentation is a combination of object detection and semantic segmentation. It requires not only objects of various sizes in an image be correctly detected but also each object be precisely segmented. For the instance segmentation task, our experiments are implemented by using the Mask R-CNN on the MS COCO dataset. As illustrated by Table \ref{tab:instance_segmantation}, the proposed EMBANet-M(Large)-50 outperforms the SENet by approximately 1.7$\%$ and 1.6$\%$ on $AP$, $AP_{50}$ respectively. The proposed EMBANet-S(Large) also achieves a large margin of 1.1 $\%$, 1.0 $\%$ and 0.5 $\%$ on $AP$ as compared to ECANet, GCNet and FcaNet respectively. The visualization results are further illustrated in Fig. \ref{fig:coco_embanet_s} and \ref{fig:coco_embanet_m}. In summary, test results on the instance segmentation illustrated the strong migration ability of our proposed architecture.

\CheckRmv{
	\begin{figure}[t]
		\centering
		\begin{overpic}[width=0.8\linewidth]{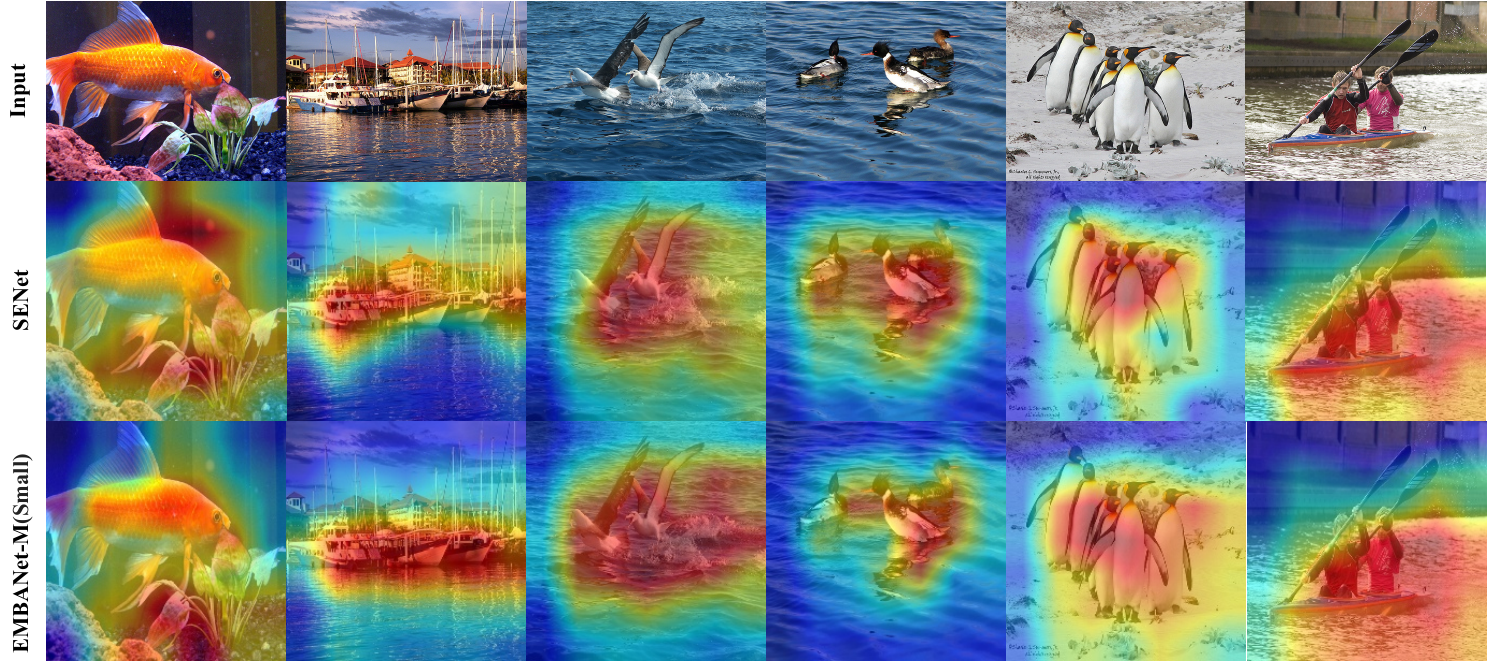}
		\end{overpic}
		\caption{Visualization of GradCAM results which are obtained for six random samples from the ImageNet validation set by using the SENet-50 and EMBANet-M(Small)-50. }
		\label{fig:cam}
	\end{figure}
}
\section{Class Activation Mapping}  
Because the MUC-based network can offer better performance for various computer vision tasks, we employed the EMBANet-M(Small) as the baseline in this experiment to demonstrate the multi-scale ability of the MBC module. Specifically, we visualized the class activation mapping (CAM) of the EMBANet-M(Small) by using Grad-CAM~\cite{selvaraju2017grad}. The Grad-CAM is commonly used to localize discriminatory regions for image classification and can also be used to visualize the gradients of top-class prediction with respect to the input image as a colored overlay. As illustrated by Fig. \ref{fig:cam}, the proposed EMBANet-M(Small) has more concentrated activation maps on small and medium objects and can generate fine-level feature representations. For intensive objects, the \ourM-M(Small) has more accurate positioning and division than that of the SENet. The visualization results verified that the MBC-based network can capture richer and more discriminatory contextual information for a particular target class.

\section{Conclusion and Future Work}
A novel multi-scale feature extraction framework, namely Multi-branch and Concat (MBC), is proposed in this work. The proposed MBC brings extra DoF and thus makes the attention network design more flexible and scalable. To implement the MBC, two important variations, the Multiplex and Concat (MUC) and the Split and Concat (SPC), are developed. Both can bring performance gains from different perspectives across various computer vision tasks. The MUC, which can achieve state-of-the-art performance in tasks of image classification and object detection, is more suitable for a large backbone. The SPC, which is more efficient as compared to existing attention methods, is more suitable for a mobile network. The proposed MBC can also learn a richer multi-scale feature representation of the input tensor at a more granular level. Correspondingly, a unified multi-scale attention network architecture called EMBANet has been further developed based on the MBC module. Extensive experiments on tasks of image classification, object detection, and image segmentation demonstrate that the proposed EMBANet can outperform CNN- and Attention-based models in terms of effectiveness and efficiency. In the future, we will design a novel search method to automatically find the optimal DoF for attention modules and detailed structures.

\section*{Acknowledgment}
This work was supported by the National Natural Science Foundation of China under grants 61972265 and 11871348, the Natural Science Foundation of Guangdong Province of China under grant 2020B1515310008, and the Educational Commission of Guangdong Province of China under grant 2019KZDZX1007, the Pazhou Lab, Guangzhou, China.

\bibliographystyle{elsarticle-num} 
\bibliography{reference}


\end{document}